\documentclass[11pt,a4paper]{article}
\usepackage[hyperref]{acl2020}
\usepackage{times}
\usepackage{latexsym}

\usepackage{microtype}

\usepackage{mathtools}
\usepackage{microtype}
\usepackage{booktabs}
\usepackage{enumitem}
\setlist{nosep}
\usepackage{cleveref}
\Crefname{equation}{Eq.}{Eqs.}
\Crefname{figure}{Fig.}{Figs.}
\Crefname{table}{Tab.}{Tabs.}
\Crefname{section}{Sec.}{Secs.}
\usepackage{subcaption}

\usepackage{graphicx}
\usepackage[normalem]{ulem}
\usepackage{todonotes}
\usepackage{float}
\usepackage[super]{nth}

\aclfinalcopy 
\def\aclpaperid{2616} 

\title{Evaluating Neural Machine Comprehension Model \\
    Robustness to Noisy Inputs and Adversarial Attacks}

\author{Winston Wu\\
  Center for Language and Speech Processing \\
  Department of Computer Science\\
  Johns Hopkins University \\
  \texttt{wswu@jhu.edu} \\\And
  Dustin Arendt\textsuperscript{1} \quad Svitlana Volkova\textsuperscript{2} \\
  \textsuperscript{1}Visual Analytics Group\\
  \textsuperscript{2}Data Sciences and Analytics Group\\
  Pacific Northwest National Laboratory \\
  \texttt{first.last@pnnl.gov}}

\date{}

\begin{document}
\maketitle
\begin{abstract}

We evaluate machine comprehension models' robustness to noise and adversarial attacks by performing novel perturbations at the character, word, and sentence level. We experiment with different amounts of perturbations to examine model confidence and misclassification rate, and contrast model performance in adversarial training with different embedding types on two benchmark datasets. We demonstrate improving model performance with ensembling. Finally, we analyze factors that effect model behavior under adversarial training and develop a model to predict model errors during adversarial attacks.
\end{abstract}

\section{Introduction}

Deep neural models have recently gained popularity, leading to significant improvements in many language understanding and modeling tasks, including machine translation, language modeling, syntactic parsing, and machine comprehension \citep{goldberg2017neural}. However, the NLP community still lacks in-depth understanding of how these models work and what kind of linguistic information is actually captured by neural networks \citep{feng2018pathologies}. Evaluating model robustness to manipulated inputs and analyzing model behavior during adversarial attacks can provide deeper insights into how much language understanding neural models actually have. Moreover, as has been widely discussed, neural models should be optimized not only for task performance but also for other important criteria such as reliability, fairness, explainability, and interpretability \citep{lipton2016mythos,doshi2017towards,ribeiro2016should,goodman2017european,hohman2018visual}.

In this work, we evaluate neural models' robustness on machine comprehension (MC), a task designed to measure a system's understanding of text. In this task, given a \textit{context} paragraph and a \textit{question}, the machine is tasked to provide an \textit{answer}. We focus on \textit{span-based} MC, where the model selects a single contiguous span of tokens in the context as the answer (\Cref{tab:mc-example}). We quantitatively measure when and how the model is robust to manipulated inputs, when it generalizes well, and when it is less susceptible to adversarial attacks.

\begin{table}[t!]
    \small
    \centering
    \begin{tabular}{p{7cm}}
        \toprule
        \textbf{Context}: One of the most famous people born in Warsaw was Maria Sk\l odowska-Curie, who achieved international recognition for her research on radioactivity and was the first female recipient of the \underline{Nobel Prize}. \\
        \textbf{Question}: What was Maria Curie the first female recipient of? \\
        \textbf{Answer}: Nobel Prize \\
        \bottomrule
    \end{tabular}
    \caption{Example MC question from SQuAD.}
    \label{tab:mc-example}
    \vspace{-0.5cm}
\end{table}

Our novel contributions shed light on the following research questions:
\begin{itemize}
    \item Which embeddings are more susceptible to noise and adversarial attacks?
    \item What types of text perturbation lead to the most high-confidence misclassifications?
    \item How does the amount of text perturbation effect model behavior? 
    \item What factors explain model behavior under perturbation?
    \item Are the above dataset-specific?
\end{itemize}


\begin{table*}[t!]
    \small
    \centering
    \begin{tabular}{l p{13cm}}
        \toprule
        Original & The connection between macroscopic nonconservative forces and microscopic conservative forces is described by detailed treatment with statistical mechanics. \\
        \midrule
        Character & \textbf{T}h\textbf{e} \textbf{c}onn\textbf{e}ct\textbf{i}\textbf{o}n b\textbf{e}twe\textbf{e}n macr\textbf{o}scopi\textbf{c} nonc\textbf{o}nservat\textbf{i}v\textbf{e} force\textbf{s} an\textbf{d} micro\textbf{s}co\textbf{p}i\textbf{c} con\textbf{s}ervative f\textbf{o}r\textbf{c}es is de\textbf{s}crib\textbf{e}d by detailed tre\textbf{a}tm\textbf{e}nt with st\textbf{a}t\textbf{i}st\textbf{i}cal m\textbf{e}\textbf{c}hanic\textbf{s}. \\
        \midrule
        Word & The connection between macroscopic nonconservative forces and \textbf{insects} conservative \textbf{troops} is \textbf{referred} by detailed treatment with statistical mechanics. \\
        \midrule
        Sentence & The \textbf{link} between macroscopic \textbf{non-conservative} forces and microscopic conservative forces is described \textbf{in detail by} statistical mechanics. \\
        \bottomrule
    \end{tabular}
    \caption{Examples of character, word and sentence-level perturbations (bold indicates perturbed text).}
    \label{tab:perturb-examples}
    \vspace{-0.3cm}
\end{table*}

\section{Background}

There is much recent work on adversarial NLP, surveyed in \citet{belinkov2018survey,zhang2019survey}. To situate our work, we review relevant research on the black-box adversarial setting, in which one does not have access or information about the model's internals, only the model's output and its confidence about the answer. 

In an adversarial setting, the adversary seeks to mislead the model into producing an incorrect output by slightly tweaking the input. Recent work has explored input perturbations at different linguistic levels: character, word, and sentence-level.

For \textit{character-level perturbations},
researchers have explored the effects of adding noise by randomizing or swapping characters and examining its effect on machine translation (MT) \citep{heigold2018robust,belinkov2017synthetic}, sentiment analysis and spam detection \citet{gao2018deepwordbug}, and toxic content detection \citet{li2018textbugger}. \citet{eger2019viper} replaced with similar looking symbols, and developed a system to replace characters with nearest neighbors in visual embedding space.
For \textit{word-level perturbations}, \citet{alzantot2018generating} used a genetic algorithm to replace words with contextually similar words, evaluating on sentiment analysis and textual entailment.
For \textit{sentence-level perturbations}, \citet{iyyer2018syntactic} generated adversarial paraphrases by controlling the syntax of sentences and evaluating on sentiment analysis and textual entailment tasks. \citet{hu2019improvedparabank} found that augmenting the training data with paraphrases can improve performance on natural language inference, question answering, and MT. \citet{niu2018adversarial} use adversarial paraphrases for dialog models.

Other related work includes \citet{zhao2017generating}, who generated natural looking adversarial examples for image classification, textual entailment, and MT. Specifically for MC, \citet{jia2017adversarial} added a distractor sentence to the end of the context, and \citet{ribeiro2018sears} extracted sentence perturbation rules from paraphrases created by translating to and then from a foreign language and then manually judged for semantic equivalence. They experimented on MC, visual question answering, and sentiment analysis.

Unlike earlier work, we empirically show how neural model performance degrades under multiple types of adversarial attacks by varying the amount of perturbation, the type of perturbation, model architecture and embeddings type, and the dataset used for evaluation. Moreover, our deep analysis examines factors that can explain neural model behavior under these different types of attacks. 

\section{Experiments}

We perform comprehensive adversarial training and model evaluation for machine comprehension over several dimensions: the amount of perturbation, perturbation type, model and embedding variation, and datasets.

\subsection{Perturbation Type} We examine how changes to the context paragraph (excluding the answer span) affect the model's performance using the following perturbations:

\begin{itemize}

\item \textbf{Character-level}. In computer security, this is known as a homograph attack. These attacks attacks have been investigated to identify phishing and spam \citep{fu2006safeguard,fu2006methodology,liu2007fighting} but to our knowledge have not been applied in the NLP domain. We replace 25\% of characters in the context paragraph with deceptive Unicode characters\footnote{From \url{https://www.unicode.org/Public/security/12.1.0/intentional.txt}} that to a human are indistinguishable from the original.
    
\item \textbf{Word-level}. We randomly replace 25\% of the words in the context paragraph with their nearest neighbor in the GLoVe \citep{pennington2014glove} embedding space.\footnote{Several alternative embedding techniques could be used to find the nearest neighbors e.g., Word2Vec or FastText. We use GLoVe for consistency with previous work \citep{li2018textbugger}.}


\item \textbf{Sentence-level}. We use Improved ParaBank Rewriter \citep{hu2019improvedparabank}, a machine translation approach for sentence paraphrasing, to paraphrase sentences in the context paragraph. We perform sentence tokenization, paraphrase each sentence with the paraphraser, then recombine the sentences. 

\end{itemize}

For character and word perturbations, we use 25\% as this is where the performance curve in \citet{heigold2018robust} flattens out.\footnote{\citet{belinkov2017synthetic} perturbed text at 100\% (every word), while \citet{heigold2018robust} experimented with 5--30\% perturbations.} Regardless of the type of perturbation, we do not perturb the context that contains the answer span, so that the answer can always be found in the context unperturbed. Because paraphrasing is per sentence, we only modify sentences that do not contain the answer span. An example of each perturbation type is shown in \Cref{tab:perturb-examples}.

\subsection{Amount of Perturbation}

For each perturbation type, we experiment with perturbing the training data at differing amounts. All models are tested on fully perturbed test data.

\begin{itemize}
    \item \textbf{None}: clean training data
    \item \textbf{Half}: perturb half the training examples
    \item \textbf{Full}: perturb the entire train set
    \item \textbf{Both}: append the entire perturbed data to the entire clean data\footnote{This has twice the amount of data as other settings so is not directly comparable, but many papers show that doing this can improve a model's performance.}
    \item \textbf{Ens}: ensemble model that relies on none, half and full perturbed data; we rely on ensemble voting and only include the word in the predicted answer if any two models agree.
\end{itemize}


\begin{figure*}[t!]
    \centering
    \begin{subfigure}[b]{0.31\textwidth}
        \includegraphics[width=\linewidth]{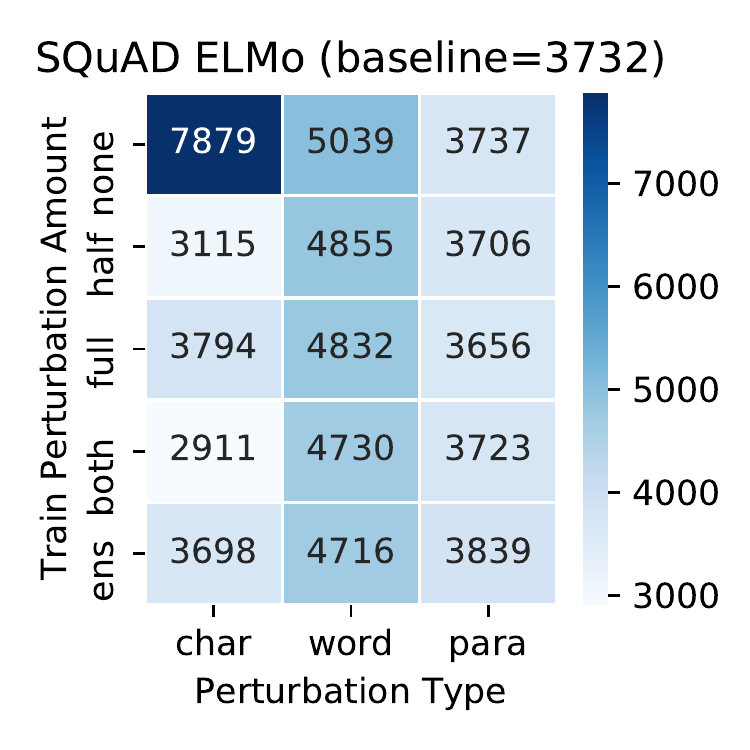}
        \caption{SQuAD, ELMo}
        \label{fig:errors-squad-elmo}
    \end{subfigure}
    ~
    \begin{subfigure}[b]{0.31\textwidth}
        \includegraphics[width=\linewidth]{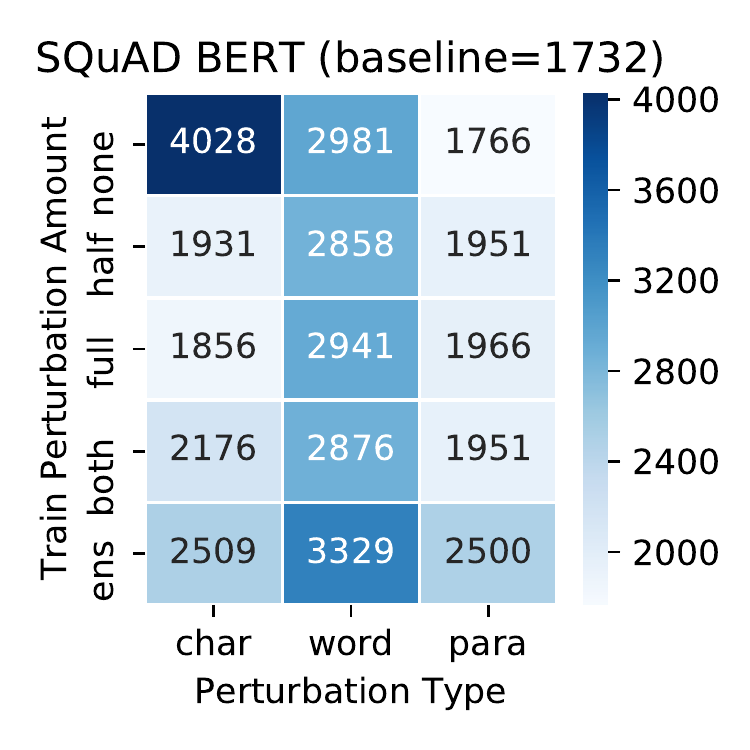}
        \caption{SQuAD, BERT}
        \label{fig:errors-squad-bert}
    \end{subfigure}
    ~
    \begin{subfigure}[b]{0.31\textwidth}
        \includegraphics[width=\linewidth]{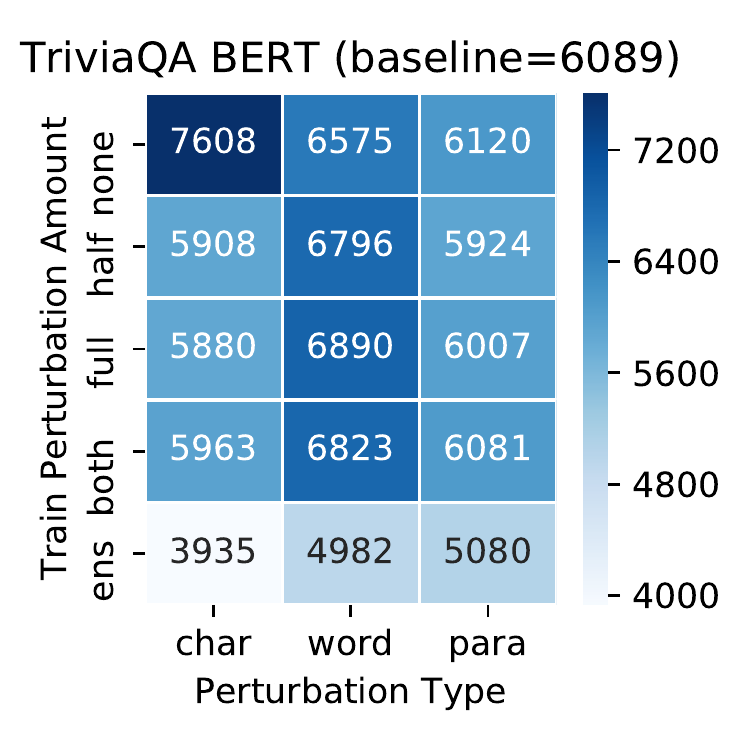}
        \caption{TriviaQA, BERT}
        \label{fig:errors-triviaqa-bert}
    \end{subfigure}
    \caption{Number of errors by perturbation type and amount of perturbation (higher = worse model performance, or more successful attacks). \textit{Baseline} indicates model errors whose training and testing data were not perturbed. For cross-model/embedding comparison, compare (a) and (b). For cross-dataset comparison, compare (a) and (c). The \textit{ens} training setting is an ensemble of results from the none, half, and full settings.}
    \label{fig:errors-summary}
\end{figure*}

\subsection{Model and Embedding}

We experiment with several recent models and embeddings that have been state-of-the-art on machine comprehension.

\begin{itemize}

\item \textbf{BiDAF model with ELMo} \cite{seo2016bidaf,peters2018elmo}. ELMo is a deep, contextualized, character-based word embedding method using a bidirectional language model. The Bi-Directional Attention Flow model is a hierarchical model with embeddings at multiple levels of granularity: character, word, and paragraph. We use pre-trained ELMo embeddings in the BiDAF model implemented in AllenNLP \citep{gardner2018allennlp}.

\item \textbf{BERT} \citep{devlin2019bert}. BERT is another contextualized embedding method that uses Transformers \citep{vaswani2017attention}. It is trained to recover masked words in a sentence as well as on a next-sentence prediction task. The output layer of BERT is fed into a fully-connected layer for the span classification task. Pre-trained embeddings can be fine-tuned to a specific task, and we use the Huggingface PyTorch-Transformers package, specifically {\it bert-large-cased-whole-word-masking-finetuned-squad} model. We fine-tune for two epochs in each experimental settings.


\end{itemize}

\subsection{Benchmark Datasets}
\label{sec:datasets}

We experiment on two benchmark MC datasets:

\begin{itemize}
    
\item \textbf{SQuAD} \citep{rajpurkar2016squad}. The Stanford Question Answering Dataset is a collection of over 100K crowdsourced question and answer pairs. The context containing the answer is taken from Wikipedia articles.

\item \textbf{TriviaQA} \citep{joshi2017triviaqa}. A collection of over 650K crowdsourced question and answer pairs, where the context is from web data or Wikipedia. The construction of the dataset differs from SQuAD in that question answer pairs were first constructed, then evidence was found to support the answer. We utilize the Wikipedia portion of TriviaQA, whose size is comparable to SQuAD. To match the span-based setting of SQuAD, we convert TriviaQA to the SQuAD format using the scripts in the official repo and remove answers without evidence.


\end{itemize}

\section{Results}

\begin{figure*}[t!]
    \centering
    \begin{subfigure}[b]{0.48\linewidth}
        \includegraphics[width=\linewidth]{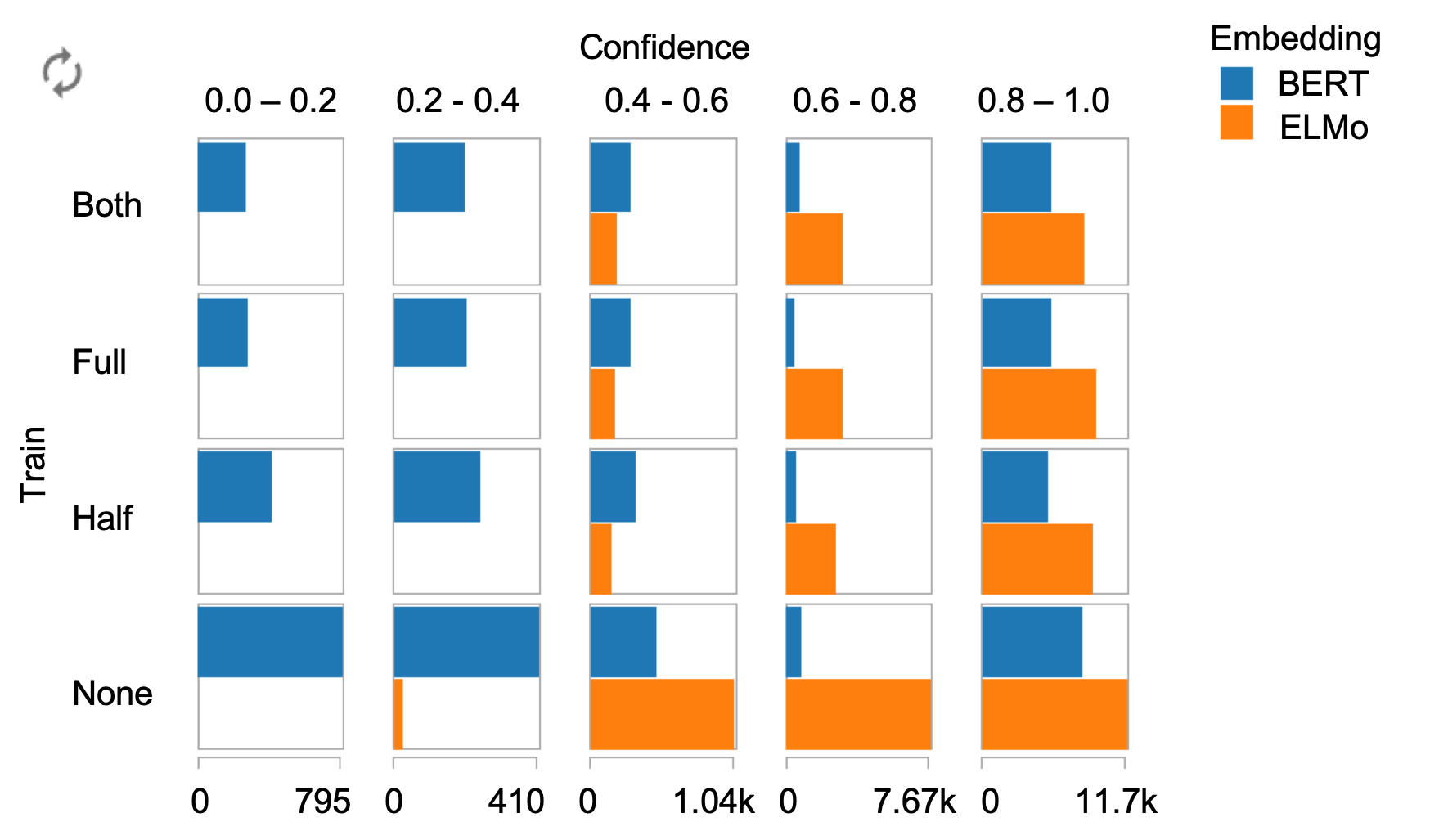}
        \caption{Across models and embedding}
        \label{fig:confidence-embedding}
    \end{subfigure}
    ~
    \begin{subfigure}[b]{0.49\linewidth}
        \includegraphics[width=\textwidth]{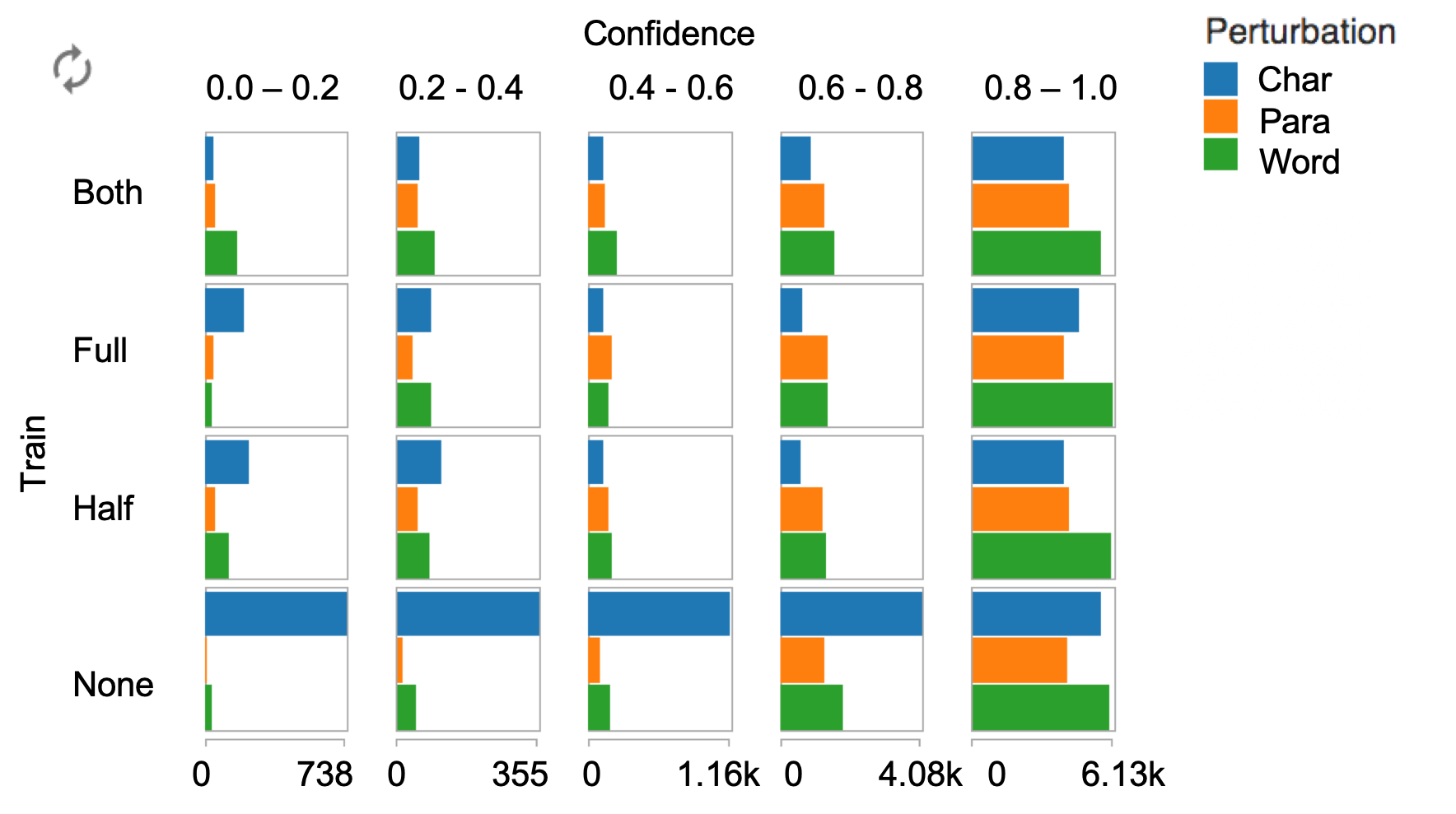}
        \caption{Across perturbation types}
        \label{fig:confidence-type}
    \end{subfigure}
    \caption{The effect of perturbation types and embeddings on model behavior measured as high vs. low confidence misclassifications under adversarial attacks. More robust models should have less high-confidence misclassifications.}
    \vspace{-0.2cm}
\end{figure*}

\Cref{fig:errors-summary} summarizes our findings on how model behavior changes under noisy perturbations and adversarial attacks. Here, we briefly discuss how perturbation type, perturbation amount, model, and embeddings affect model misclassification rate. In addition, we contrast model performance across datasets and report how to mitigate model error rate using ensembling. Detailed analyses are presented in \Cref{sec:analysis}. Key findings are \textit{italicized}. 

\paragraph{Perturbation type} To assess whether perturbations changed the meaning of the context, we ran a human study on a random sample of 100 perturbed contexts from SQuAD. We found (as expected) that humans could not distinguish character-perturbed text from the original. For word perturbations, the meaning of the context remained in 65\% of cases, but annotators noted that sentences were often ungrammatical. For sentence-level perturbations, the meaning remained the same in 83\% of cases.

\textit{For a model trained on clean data, character perturbations affect the model the most, followed by word perturbations, then paraphrases.} To a machine, a single character perturbation results in a completely different word; handling this type of noise is important for a machine seeking to beat human performance. Word perturbations are context independent and can make the sentence ungrammatical.\footnote{Future work will address this with language models.} Nevertheless, the context's meaning generally remains coherent. Paraphrase perturbations are most ideal because they retain meaning while allowing more drastic phrase and sentence structure modifications. In \Cref{sec:strategic-paraphrasing}, we present a more successful adversarially targeted paraphrasing approach.

\paragraph{Perturbation amount} Perturbed training data improves the model's performance for character perturbations (\nth{1} column of \Cref{fig:errors-summary}a), likely due to the models' ability to handle unseen words: BiDAF with ELMo utilizes character embeddings, while BERT uses word pieces. Our results corroborate \citet{heigold2018robust}'s findings (though on a different task) that \textit{without adversarial training, models perform poorly on perturbed test data, but when models are trained on perturbed data, the amount of perturbed training data does not make much difference.} We do not see statistically significant results for word and paraphrase perturbations (\nth{2} and \nth{3} columns in each heatmap in \Cref{fig:errors-summary}). We conclude that perturbing 25\% of the words and the non-strategic paraphrasing approach were not aggressive enough.

\paragraph{Model and embedding} As shown in \Cref{fig:errors-summary}a and b, the BERT model had less errors than the ELMO-based model regardless of the perturbation type and amount on SQuAD data. While the two models are not directly comparable, our results indicate that the \textit{BERT model is more robust to adversarial attacks compared to ELMo.}

\paragraph{Datasets}  Holding the model constant (\Cref{fig:errors-summary}b and c), experiments on TriviaQA resulted in more errors than SQuAD regardless of perturbation amount and type, indicating that TriviaQA may be a harder dataset for MC and may contain data bias, discussed below.

\begin{table}[t!]
    \small
    \centering
    \begin{tabular}{lll}
        \toprule
        Train & Test & Model Answer \\
        \midrule
        none & char & here'' \\
        half & char & Orientalism \\
        full & char & Orientalism \\
        \midrule
        none & word & Orientalism \\
        half & word & behaviourism identities \\
        full & word & The discourse of Orientalism \\
        \midrule
        none & char & Orientalism \\
        half & char & \dots the East as a negative\\
        full & char & Orientalism \\
        \bottomrule
    \end{tabular}
    \caption{Example result from response ensembling under the SQuAD ELMo setting. The question is ``What was used by the West to justify control over eastern territories?'' The answer is ``Orientalism'', and in all three settings, the ensemble was correct.}
    \label{tab:ensemble}
    \vspace{-0.3cm}
\end{table}

\newcommand{\constraint}[1]{\textcolor{red}{#1}}
\newcommand{\changed}[1]{\textcolor{blue}{#1}}
\newcommand{\correct}[1]{\textcolor{Green}{#1}}

\begin{table*}[t!]
    \small
    \centering
    \begin{tabular}{p{7cm} p{7cm}}
        \toprule
        Original Paragraph & Strategic Paraphrase\\
        \midrule
        $\ldots$ \constraint{Veteran} \constraint{receiver} Demaryius Thomas \constraint{led} the team with \constraint{105} receptions for 1,304 yards and six touchdowns, while Emmanuel Sanders caught 76 passes for 1,135 yards and six scores, while adding another 106 yards returning punts.
        & $\ldots$ \changed{The veteran earman} Demaryius Thomas \changed{was leading a} team of 1,304 yards and six touchdowns, while Emmanuel Sanders caught 76 passes for 1,135 yards and six scores while \changed{he added} another 106 yards \changed{of punts back}. \\
        \midrule
        \multicolumn{2}{l}{
            \textbf{Question}: Who led the Broncos with 105 receptions?
        } \\
        \multicolumn{2}{l}{
            \textbf{Answer}: Demaryius Thomas (correct) $\to$ Emmanuel Sanders (incorrect)
        } \\
        \bottomrule
    \end{tabular}
    \caption{Example of strategic paraphrasing: red indicates the important words, which were used as negative constraints in the paraphrasing; blue indicates changed words in the paragraph.}
    \label{tab:strategic-paraphrase}
\end{table*}

\subsection{Adversarial Ensembles}

Ensemble adversarial training has recently been explored \citep{tramer2017ensemble} as a way to ensure robustness of ML models. For each perturbation type, we present results ensembled from the none, half, and full perturbed settings. We tokenize answers from these three models and keep all tokens that appear at least twice as the resulting answer (\Cref{tab:ensemble}). Even when all three model answers differ (e.g. in the word perturbation case), ensembling can often reconstruct the correct answer. Nevertheless, we find that this ensembling only helps for TriviaQA, which has an overall higher error rate (bottom row of each figure in \Cref{fig:errors-summary}).

\subsection{Strategic Paraphrasing}
\label{sec:strategic-paraphrasing}

We did not observe a large increase in errors with paraphrase perturbations (\Cref{fig:errors-summary}), perhaps because paraphrasing, unlike the char and word perturbations, is not a deliberate attack on the sentence. Here we experiment with a novel strategic paraphrasing technique that targets specific words in the context and then generates paraphrases that exclude those words. We find the most important words in the context by individually modifying each word and obtaining the model's prediction and confidence, a process similar to \citet{li2018textbugger}. Our modification consists of removing the word and examining its effect on the model prediction. The most important words are those which, when removed, lower the model confidence of a correct answer or increase confidence of an incorrect answer. The Improved ParaBank Rewriter supports constrained decoding, i.e. specifying positive and negative constraints to force the system output to include or exclude certain phrases. We specify the top five important words in the context as negative constraints to generate strategic paraphrases.\footnote{The number of constraints does not necessarily indicate the number of words that are changed in the context.} 

We experimented on 1000 instances in the SQuAD dev set as shown in \Cref{tab:strategic-paraphrase}.
Our results indicate that \textit{strategic paraphrasing with negative constraints is a successful adversarial attack}, lowering the F1-score from 89.96 to 84.55.

Analysis shows that many words in the question are important and thus excluded from the paraphrases. We also notice that paraphrasing can occasionally turn an incorrect prediction into a correct one. Perhaps paraphrasing makes the context easier to understand by removing distractor terms; we leave this for future investigation.

\subsection{Model Confidence}

In a black-box setting, model confidence is one of the only indications of the model's inner workings. The models we employed do not provide a single confidence value; AllenNLP gives a probability that each word in the context is the start and end span, while the BERT models only give the probability for the start and end words. We compute the model's confidence using the normalized entropy
of the distribution across the context words, where $n$ is the number of context words, and take the mean for both the start and end word:
$ 1 - \frac{ H_{n}(\mathbf{s}) + H_{n}(\mathbf{e}) }{ 2 } $,
where $\mathbf{s}$ and $\mathbf{e}$ are probability distributions for the start and end words, respectively. Low entropy indicates certainty about the start/end location. Since the BERT models only provide probabilities for the start and end words, we approximate the entropy by assuming a flat distribution, dividing the remaining probability equally across all other words in the context.

Comparing confidence across models (\Cref{fig:confidence-embedding}), the \textit{BERT model has lower confidence for misclassifications}, which is ideal. A model should not be confident about errors. \Cref{fig:confidence-type} compares confidence across perturbation type. In the \textit{none} training setting, character perturbations introduce the most uncertainty compared to word or paraphrase perturbations. This is expected, since character perturbations result in unknown words. In the adversarial training settings, word perturbations lead to the highest number of high-confidence misclassifications. Thus, \textit{to convincingly mislead the model to be highly confident about errors, one should use word perturbations.}

\section{Robustness Analysis}
\label{sec:analysis}

Here, we do a deeper dive into why models make errors with noisy input. We investigate data characteristics and their association with model errors by utilizing CrossCheck \citep{arendt2020crosscheck}, a novel interactive tool designed for neural model evaluation. Unlike several recently developed tools for analyzing NLP model errors \cite{agarwal2014error,wu2019errudite} and understanding ML model outputs \cite{lee2019qadiver,poursabzi2018manipulating,hohman2019gamut}, CrossCheck is designed to allow rapid prototyping and cross-model comparison to support comprehensive experimental setup. CrossCheck interfaces with Jupyter notebooks and supports custom interactive widgets such as heatmaps and histograms grids.

\begin{figure*}[t!]
    \centering
    \begin{subfigure}[t]{0.29\linewidth}
        \includegraphics[width=\linewidth]{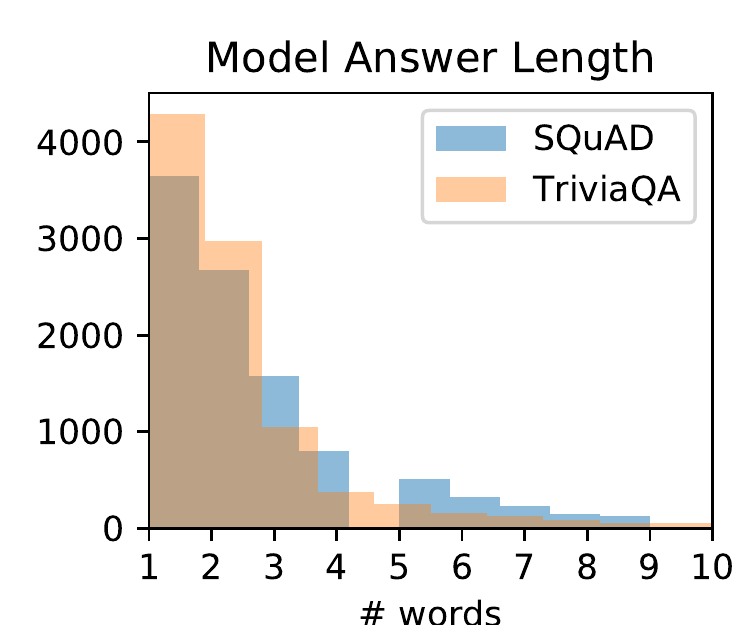}
    \end{subfigure}
    ~
    \begin{subfigure}[t]{0.33\linewidth}
        \includegraphics[width=\linewidth]{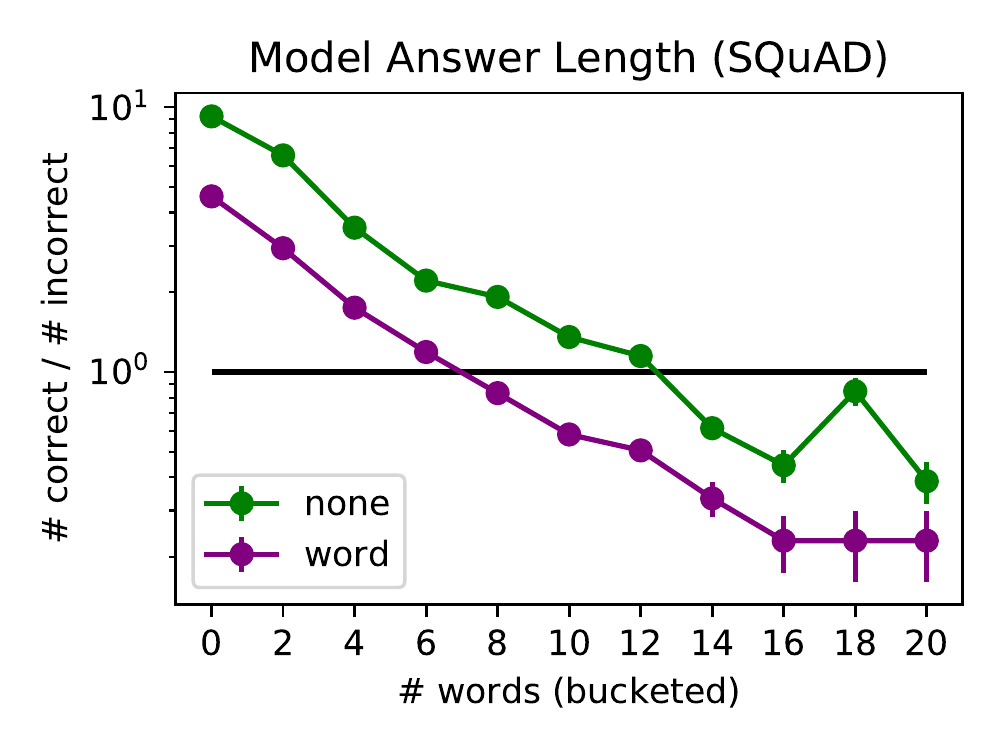}
    \end{subfigure}
    ~
    \begin{subfigure}[t]{0.33\linewidth}
        \includegraphics[width=\linewidth]{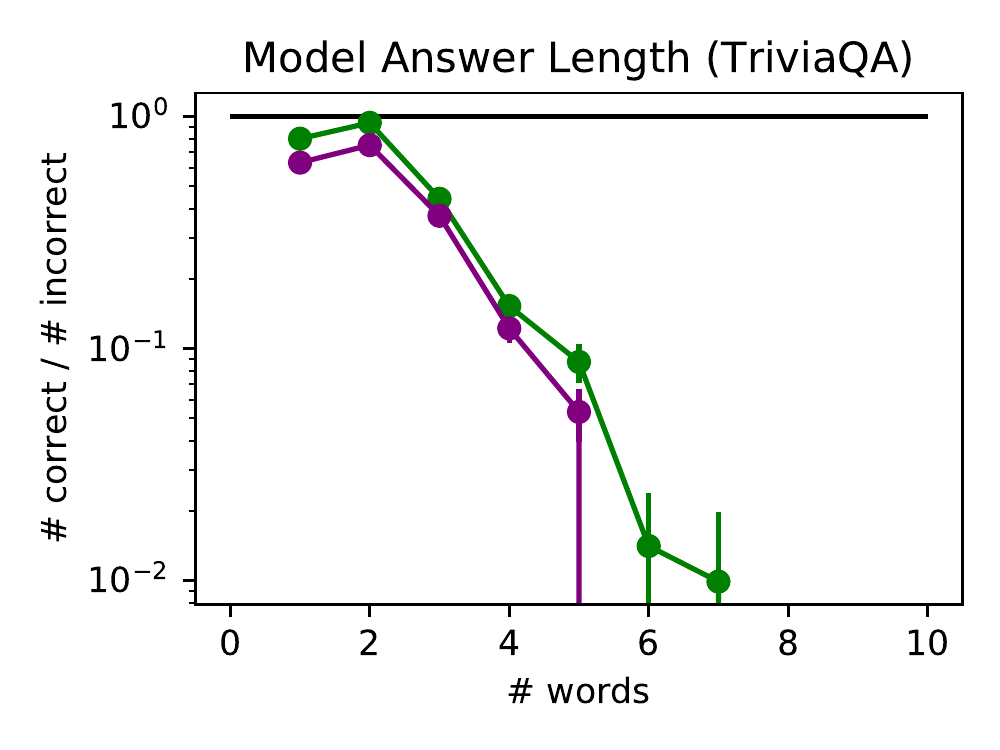}
    \end{subfigure}
    \caption{(a) shows distribution of answer length. (b) and (c) show the ratio of correct errors to incorrect errors (log scale y-axis). BERT models were trained on clean data and tested on perturbed data. Standard error was used for the error bar amounts.}
    \label{fig:ratios}
\end{figure*}

\subsection{The Effect of Question Type, Question and Context Lengths}

\begin{figure*}[t!]
    \centering
    \begin{subfigure}[t]{0.3\linewidth}
        \includegraphics[width=\linewidth]{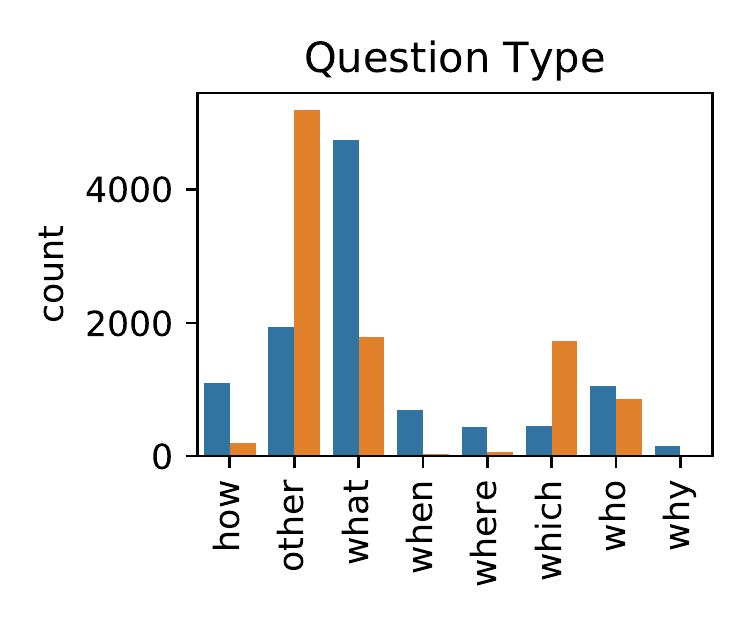}
        \caption{Question type distribution}
        \label{fig:question-type}
    \end{subfigure}
    ~
    \begin{subfigure}[t]{0.3\linewidth}
        \includegraphics[width=\linewidth]{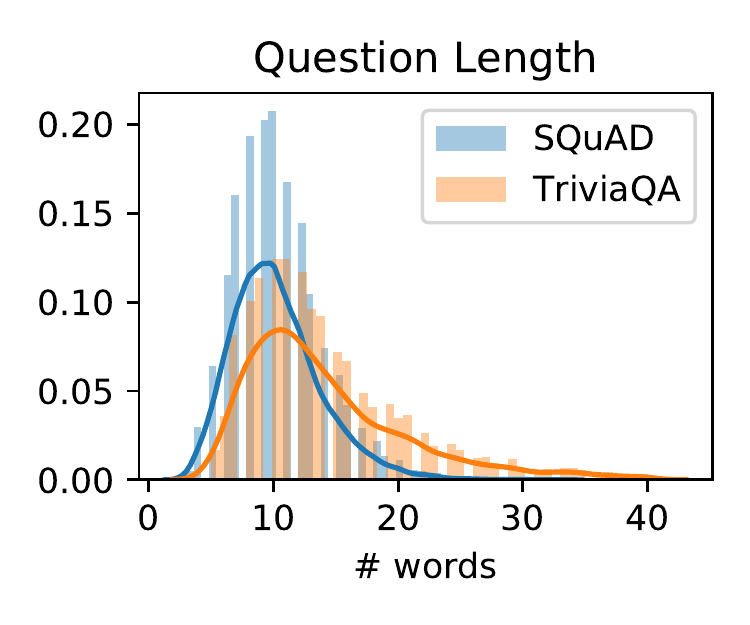}
        \caption{Question length distribution}
        \label{fig:question-length}
    \end{subfigure}
    ~
    \begin{subfigure}[t]{0.3\linewidth}
        \includegraphics[width=\linewidth]{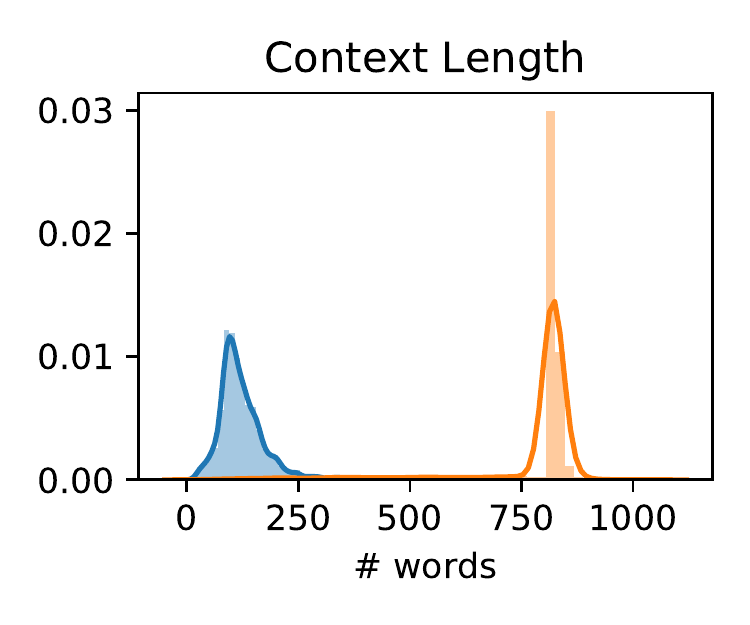}
        \caption{Context length distribution}
        \label{fig:context-length}
    \end{subfigure}
    ~
    \begin{subfigure}[t]{1.0\linewidth}
        \includegraphics[width=\linewidth]{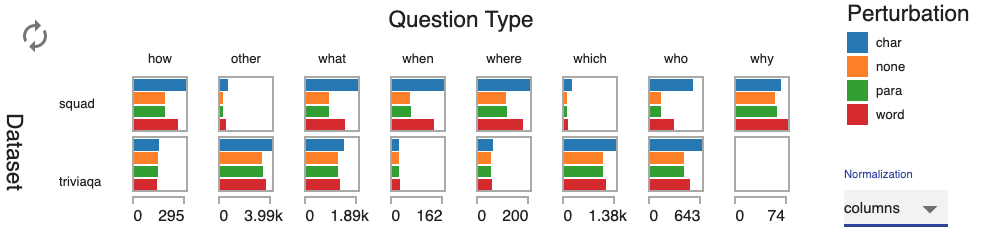}
        \caption{Error distribution in BERT models across different perturbations and question types}
        \label{fig:question-type-errors-all}
    \end{subfigure}
    
    \caption{Contrasting model errors by question type, and question and context length between SQUAD and TriviaQA datasets.}
    \label{fig:characteristics}
\end{figure*}

We examine if models make more errors on specific types of questions in adversarial training, i.e., some questions could just be easier that others. We first examine \textbf{question type}:\footnote{Computed as the first word of the question. Many \textit{how} questions are \textit{how many} or \textit{how much}, rather than how in the ``in what manner'' sense.} who, what, which, when, where, why, how, and other. The majority of SQuAD questions are \textit{what} questions, while most TriviaQA questions are \textit{other} questions, perhaps indicating more complex questions (\Cref{fig:question-type}). We see that models usually choose answers appropriate for the question type; even if they are incorrect, answers to \textit{when} questions will be dates or time word spans, and answers to \textit{how many} questions will be numbers.
\Cref{fig:question-type} presents key findings on differences in model misclassifications between two datasets given specific question types. \textit{On the SQuAD dataset, the model finds certain question types, e.g. when and how, easiest to answer regardless of the perturbation type}. Responses to these questions, which generally expect numeric answers, are not greatly affected by perturbations. For TriviaQA, in general we observe more errors across question types compared to SQuAD, i.e. more errors in what, which and who questions.

Regarding \textbf{question length}, SQuAD and TriviaQA have similar distributions (\Cref{fig:question-length}). Both datasets have a mode answer length around 10 words; TriviaQA has a slightly longer tail in the distribution. We did not find question length to impact the error. Regarding \textbf{context length}, SQuAD and TriviaQA have vastly differing context length distributions (\Cref{fig:context-length}), partly due to how the two datasets were constructed (see \Cref{sec:datasets} for details). For both datasets, the error distribution mirrors the context length distribution, and we did not find any relation between model errors and context length.

\begin{figure}[b!]
    \centering
    \vspace{-0.3cm}
    \begin{subfigure}[b]{0.48\linewidth}
        \includegraphics[width=\linewidth]{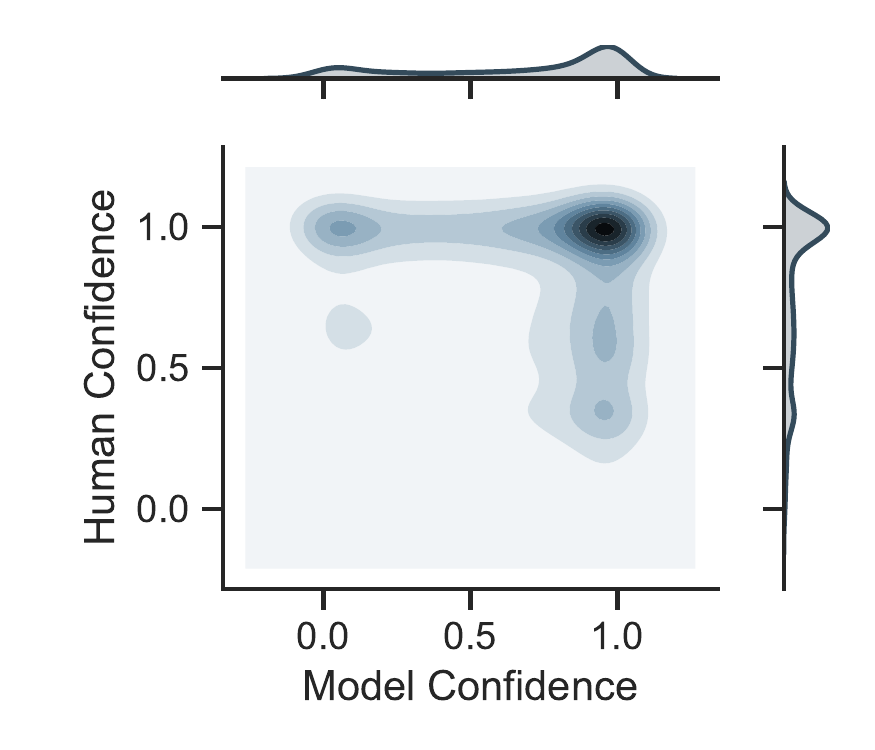}
        \caption{None perturbed}
    \end{subfigure}
    ~
    \begin{subfigure}[b]{0.48\linewidth}
        \includegraphics[width=\linewidth]{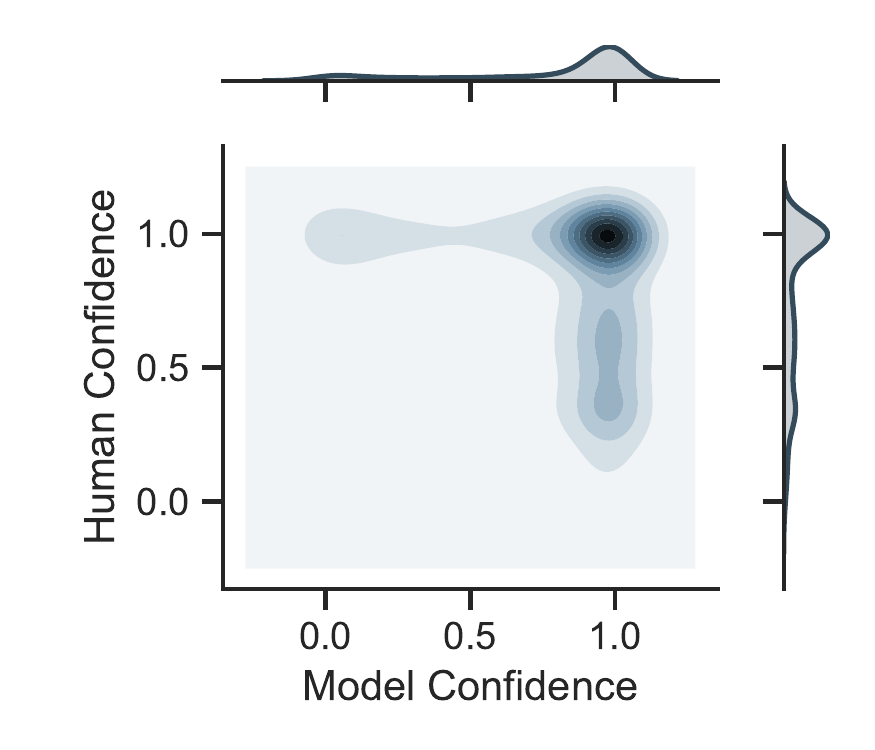}
        \caption{Half perturbed}
    \end{subfigure}
    ~\\
    \begin{subfigure}[b]{0.48\linewidth}
        \includegraphics[width=\linewidth]{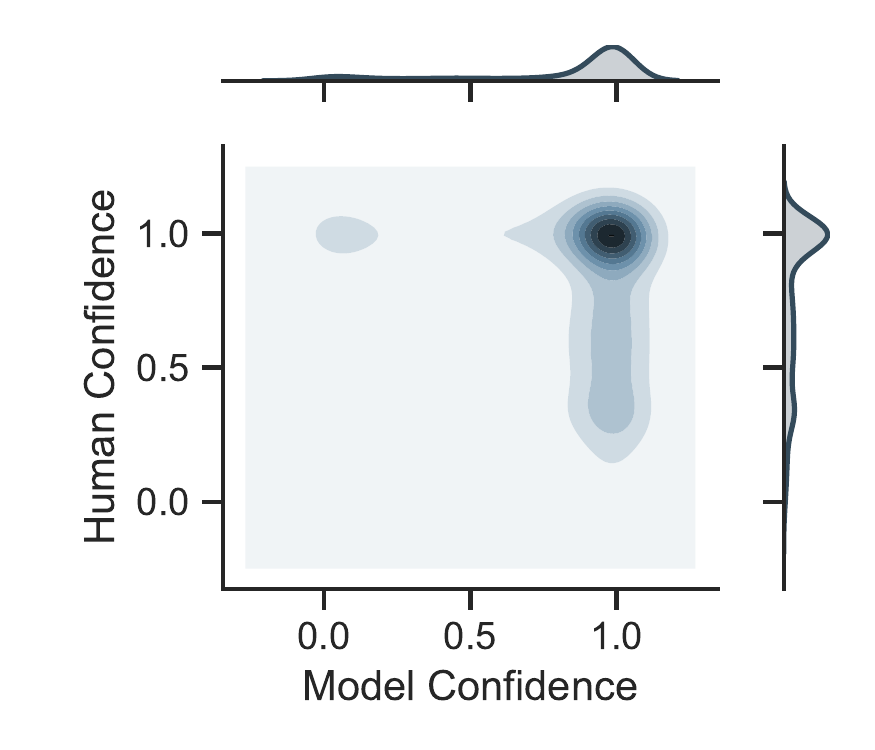}
        \caption{Full perturbed}
    \end{subfigure}
    ~
    \begin{subfigure}[b]{0.48\linewidth}
        \includegraphics[width=\linewidth]{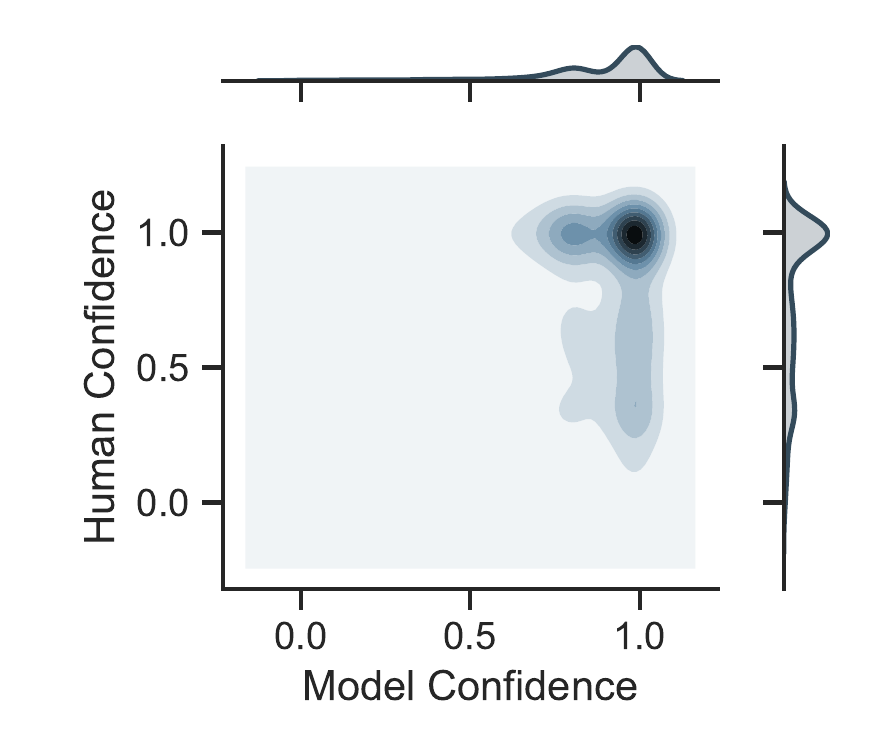}
        \caption{Both}
    \end{subfigure}
    ~
    \caption{The effect of task complexity on model behavior in adversarial setting measured as a joint distribution  of errors from BERT model on SQuAD using varied amounts of char perturbations (none, half, full and both).}
    \label{fig:confidence-annotator}
\end{figure}

\subsection{The Effect of Answer Length}

Our analysis shows that {\it the length of the model's answer is a strong predictor of model error in the adversarial setting}: the longer the answer length, the more likely it is to be incorrect. \Cref{fig:ratios} plots the proportion of correct to incorrect answers. We notice a downward trend which is mostly consistent across experimental settings. For both SQuAD and TriviaQA, the models favored shorter answers, which mirrors the data distribution.

\subsection{The Effect of Complexity: Annotator Agreement and Reading Level}

Here, we examine the effect of task complexity on model performance under adversarial training, using inter-annotator agreement as a proxy for \textit{question} complexity and paragraph readability as a proxy for \textit{context} complexity.

Inter-annotator agreement represents a \textbf{question's complexity}: low agreement indicates that annotators did not come to a consensus on the correct answer; thus the question may be difficult to answer. 
We examine SQuAD, whose questions have one to six annotated answers. In \Cref{fig:confidence-annotator}, we present inter-annotator agreement (human confidence) plotted against model confidence over the four training perturbation amounts, looking only at the incorrect predictions. The setting is SQuAD BERT with character perturbation. We observe that the models are generally confident even when the humans are not, which is noticeable across all perturbation amounts. However, we see interesting differences in model confidence in adversarial training: models trained in the none and half settings have confidence ranging between 0 and 1 compared to the models trained in full and both setting with confidence above 0.8, indicating \textit{training with more perturbed data leads to more confident models}.

To evaluate the effect of \textbf{context complexity}, we use the Flesch-Kincaid reading level \citep{kincaid1975derivation} to measure readability. For questions the model answered incorrectly, the median readability score was slightly higher than the median score for correct responses (\Cref{tab:readability}), indicating that \textit{context with higher reading level is harder for the model to understand}. TriviaQA contexts have higher reading levels than SQuAD.

\begin{table}
    \small
    \centering
    \begin{tabular}{lrr}
        \toprule
        Data   & Correct & Errors \\
        \midrule
        SQuAD & 12.9 & 13.0 \\
        TriviaQA & 17.1 & 17.5 \\
        \bottomrule
    \end{tabular}
    \caption{Contrasting median readability scores for paragraphs with and without errors across datasets. }
    \label{tab:readability}
\end{table}

\section{Error Prediction Model}

\begin{table}[t!]
    \small
    \centering
    \begin{tabular}{llll}
        \toprule
        Embedding & Pert. & Majority & F1 score\\ 
        \midrule
        ELMo & char & 0.58 & 0.70 $\pm$ 0.003\\ 
        ELMo & word & 0.54 & 0.56 $\pm$ 0.004\\
        ELMo & para & 0.65 & 0.65 $\pm$ 0.008\\
        \midrule
        BERT & char & 0.76 & 0.77 $\pm$ 0.008 \\
        BERT & word & 0.72 & 0.73 $\pm$ 0.006 \\
        BERT & para & 0.82 & 0.82 $\pm$ 0.006 \\
        \bottomrule
    \end{tabular}
    \caption{MC error prediction across both datasets, embedding, and perturbation types. Majority denotes a majority baseline.}
    \label{tab:prediction-accuracy}
\end{table}

Our in-depth analysis reveals many insights on how and why models make mistakes during adversarial training.
Using the characteristics we analyzed above, we developed a binary classification model to predict whether the answer would be an error, given the model's answer and attributes of the context paragraph. We one-hot-encode categorical features (training amount, perturbation type, question type) and use other features (question length, context length, answer length, readability) as is. For each setting of embedding and perturbation type on SQuAD, we train an XGBoost model with default settings with 10-fold cross validation (shuffled).

\begin{figure}[t!]
    \centering
  
    \includegraphics[width=\linewidth]{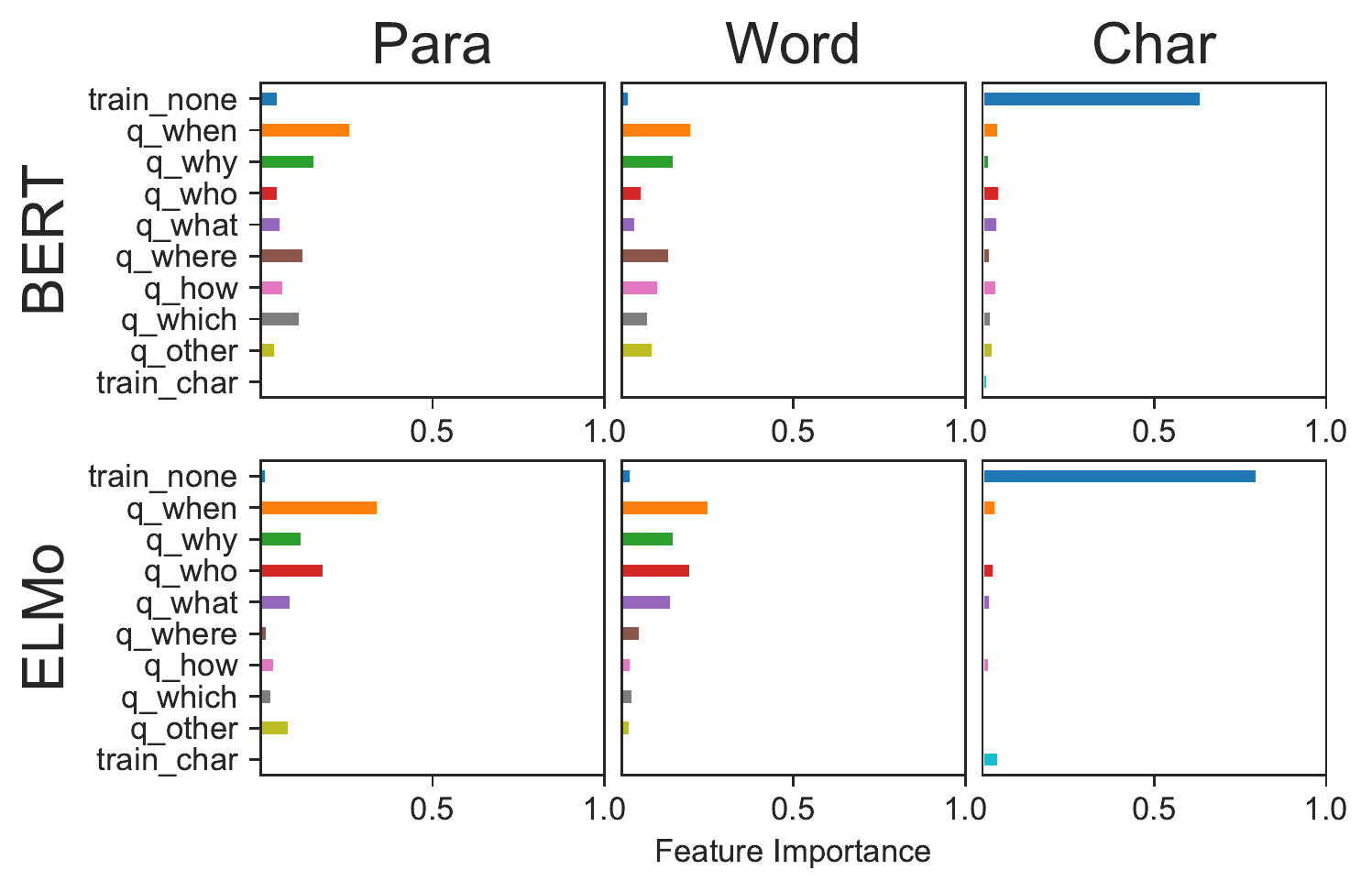}
    \caption{Feature importance when predicting model errors during adversarial attacks.}
    \label{fig:important-features}
\end{figure}

We present the model's average F1 scores (\Cref{tab:prediction-accuracy}) and feature importance as computed by the XGBoost model (\Cref{fig:important-features}). We see that performance (micro F1) is better to slightly better than a majority baseline (picking the most common class), indicating that certain features are predictive of errors. Specifically, we find that: (1) for character perturbations, the fact that the training data is clean is a strong predictor of errors; a model trained on clean data is most disrupted by character perturbations; (2) for word and paraphrase perturbations, question types are important predictors of errors.

\section{Conclusion and Future Work}

We presented a comprehensive set of experiments on machine comprehension neural models across various dimensions: perturbation type, perturbation amount, model and embedding type, and two benchmark datasets, to understand and quantitatively measure model's robustness and sensitivity to noise and adversarial input. We demonstrated that several perturbation methods, which do not greatly alter a human's comprehension of the text, can drastically affect the model's answers.
Our in-depth analysis sheds light on how and why the models make errors in adversarial training, and through our error prediction model, we discovered features of the data that are strongly predictive of when a model makes errors during adversarial attacks with noisy inputs.

For future work, we see many avenues for extension. We plan to experiment with more aggressive and more natural perturbations, and deeper counterfactual evaluation \citep{pearl2019seven}. While recent research has made great strides in increasing model performance on various NLP tasks, it is still not clear what linguistic patterns these neural models are learning, or whether they are learning language at all \citep{mudrakarta2018did}.



\bibliography{acl2020}

\begin{thebibliography}{42}
\expandafter\ifx\csname natexlab\endcsname\relax\def\natexlab#1{#1}\fi

\bibitem[{Agarwal et~al.(2014)Agarwal, Agarwal, and Mittal}]{agarwal2014error}
Apoorv Agarwal, Ankit Agarwal, and Deepak Mittal. 2014.
\newblock An error analysis tool for natural language processing and applied
  machine learning.
\newblock In \emph{Proceedings of COLING 2014, the 25th International
  Conference on Computational Linguistics: System Demonstrations}, pages 1--5.

\bibitem[{Alzantot et~al.(2018)Alzantot, Sharma, Elgohary, Ho, Srivastava, and
  Chang}]{alzantot2018generating}
Moustafa Alzantot, Yash Sharma, Ahmed Elgohary, Bo-Jhang Ho, Mani Srivastava,
  and Kai-Wei Chang. 2018.
\newblock \href {https://doi.org/10.18653/v1/D18-1316} {Generating natural
  language adversarial examples}.
\newblock In \emph{Proceedings of the 2018 Conference on Empirical Methods in
  Natural Language Processing}, pages 2890--2896, Brussels, Belgium.
  Association for Computational Linguistics.

\bibitem[{Arendt et~al.(2020)Arendt, Huang, Shrestha, Ayton, Glenski, and
  Volkova}]{arendt2020crosscheck}
Dustin Arendt, Zhuanyi Huang, Prasha Shrestha, Ellyn Ayton, Maria Glenski, and
  Svitlana Volkova. 2020.
\newblock \href {http://arxiv.org/abs/2004.07993} {Crosscheck: Rapid,
  reproducible, and interpretable model evaluation}.

\bibitem[{Belinkov and Bisk(2018)}]{belinkov2017synthetic}
Yonatan Belinkov and Yonatan Bisk. 2018.
\newblock \href {https://openreview.net/forum?id=BJ8vJebC-} {Synthetic and
  natural noise both break neural machine translation}.
\newblock In \emph{6th International Conference on Learning Representations,
  {ICLR} 2018, Vancouver, BC, Canada, April 30 - May 3, 2018, Conference Track
  Proceedings}.

\bibitem[{Belinkov and Glass(2019)}]{belinkov2018survey}
Yonatan Belinkov and James Glass. 2019.
\newblock \href {https://doi.org/10.1162/tacl_a_00254} {Analysis methods in
  neural language processing: A survey}.
\newblock \emph{Transactions of the Association for Computational Linguistics},
  7:49--72.

\bibitem[{Devlin et~al.(2019)Devlin, Chang, Lee, and
  Toutanova}]{devlin2019bert}
Jacob Devlin, Ming-Wei Chang, Kenton Lee, and Kristina Toutanova. 2019.
\newblock \href {https://doi.org/10.18653/v1/N19-1423} {{BERT}: Pre-training of
  deep bidirectional transformers for language understanding}.
\newblock In \emph{Proceedings of the 2019 Conference of the North {A}merican
  Chapter of the Association for Computational Linguistics: Human Language
  Technologies, Volume 1 (Long and Short Papers)}, pages 4171--4186,
  Minneapolis, Minnesota. Association for Computational Linguistics.

\bibitem[{Doshi-Velez and Kim(2017)}]{doshi2017towards}
Finale Doshi-Velez and Been Kim. 2017.
\newblock Towards a rigorous science of interpretable machine learning.
\newblock \emph{arXiv preprint arXiv:1702.08608}.

\bibitem[{Eger et~al.(2019)Eger, {\c{S}}ahin, R{\"u}ckl{\'e}, Lee, Schulz,
  Mesgar, Swarnkar, Simpson, and Gurevych}]{eger2019viper}
Steffen Eger, G{\"o}zde~G{\"u}l {\c{S}}ahin, Andreas R{\"u}ckl{\'e}, Ji-Ung
  Lee, Claudia Schulz, Mohsen Mesgar, Krishnkant Swarnkar, Edwin Simpson, and
  Iryna Gurevych. 2019.
\newblock \href {https://doi.org/10.18653/v1/N19-1165} {Text processing like
  humans do: Visually attacking and shielding {NLP} systems}.
\newblock In \emph{Proceedings of the 2019 Conference of the North {A}merican
  Chapter of the Association for Computational Linguistics: Human Language
  Technologies, Volume 1 (Long and Short Papers)}, pages 1634--1647,
  Minneapolis, Minnesota. Association for Computational Linguistics.

\bibitem[{Feng et~al.(2018)Feng, Wallace, Grissom~II, Iyyer, Rodriguez, and
  Boyd-Graber}]{feng2018pathologies}
Shi Feng, Eric Wallace, Alvin Grissom~II, Mohit Iyyer, Pedro Rodriguez, and
  Jordan Boyd-Graber. 2018.
\newblock \href {https://doi.org/10.18653/v1/D18-1407} {Pathologies of neural
  models make interpretations difficult}.
\newblock In \emph{Proceedings of the 2018 Conference on Empirical Methods in
  Natural Language Processing}, pages 3719--3728, Brussels, Belgium.
  Association for Computational Linguistics.

\bibitem[{Fu et~al.(2006{\natexlab{a}})Fu, Deng, Wenyin, and
  Little}]{fu2006methodology}
Anthony~Y Fu, Xiaotie Deng, Liu Wenyin, and Greg Little. 2006{\natexlab{a}}.
\newblock The methodology and an application to fight against unicode attacks.
\newblock In \emph{Proceedings of the second symposium on Usable privacy and
  security}, pages 91--101. ACM.

\bibitem[{Fu et~al.(2006{\natexlab{b}})Fu, Zhang, Deng, and
  Wenyin}]{fu2006safeguard}
Anthony~Y Fu, Wan Zhang, Xiaotie Deng, and Liu Wenyin. 2006{\natexlab{b}}.
\newblock Safeguard against unicode attacks: generation and applications of
  uc-simlist.
\newblock In \emph{WWW}, pages 917--918.

\bibitem[{Gao et~al.(2018)Gao, Lanchantin, Soffa, and Qi}]{gao2018deepwordbug}
Ji~Gao, Jack Lanchantin, Mary~Lou Soffa, and Yanjun Qi. 2018.
\newblock Black-box generation of adversarial text sequences to evade deep
  learning classifiers.
\newblock In \emph{2018 IEEE Security and Privacy Workshops (SPW)}, pages
  50--56. IEEE.

\bibitem[{Gardner et~al.(2018)Gardner, Grus, Neumann, Tafjord, Dasigi, Liu,
  Peters, Schmitz, and Zettlemoyer}]{gardner2018allennlp}
Matt Gardner, Joel Grus, Mark Neumann, Oyvind Tafjord, Pradeep Dasigi,
  Nelson~F. Liu, Matthew Peters, Michael Schmitz, and Luke Zettlemoyer. 2018.
\newblock \href {https://doi.org/10.18653/v1/W18-2501} {{A}llen{NLP}: A deep
  semantic natural language processing platform}.
\newblock In \emph{Proceedings of Workshop for {NLP} Open Source Software
  ({NLP}-{OSS})}, pages 1--6, Melbourne, Australia. Association for
  Computational Linguistics.

\bibitem[{Goldberg(2017)}]{goldberg2017neural}
Yoav Goldberg. 2017.
\newblock Neural network methods for natural language processing.
\newblock \emph{Synthesis Lectures on Human Language Technologies},
  10(1):1--309.

\bibitem[{Goodman and Flaxman(2017)}]{goodman2017european}
Bryce Goodman and Seth Flaxman. 2017.
\newblock European union regulations on algorithmic decision-making and a
  ``right to explanation''.
\newblock \emph{AI Magazine}, 38(3):50--57.

\bibitem[{Heigold et~al.(2018)Heigold, Varanasi, Neumann, and van
  Genabith}]{heigold2018robust}
Georg Heigold, Stalin Varanasi, G{\"u}nter Neumann, and Josef van Genabith.
  2018.
\newblock \href {https://www.aclweb.org/anthology/W18-1807} {How robust are
  character-based word embeddings in tagging and {MT} against wrod scramlbing
  or randdm nouse?}
\newblock In \emph{Proceedings of the 13th Conference of the Association for
  Machine Translation in the {A}mericas (Volume 1: Research Papers)}, pages
  68--80, Boston, MA. Association for Machine Translation in the Americas.

\bibitem[{Hohman et~al.(2019)Hohman, Head, Caruana, DeLine, and
  Drucker}]{hohman2019gamut}
Fred Hohman, Andrew Head, Rich Caruana, Robert DeLine, and Steven~M Drucker.
  2019.
\newblock Gamut: A design probe to understand how data scientists understand
  machine learning models.
\newblock In \emph{Proceedings of the 2019 CHI Conference on Human Factors in
  Computing Systems}, page 579. ACM.

\bibitem[{Hohman et~al.(2018)Hohman, Kahng, Pienta, and
  Chau}]{hohman2018visual}
Fred~Matthew Hohman, Minsuk Kahng, Robert Pienta, and Duen~Horng Chau. 2018.
\newblock Visual analytics in deep learning: An interrogative survey for the
  next frontiers.
\newblock \emph{IEEE transactions on visualization and computer graphics}.

\bibitem[{Hu et~al.(2019)Hu, Khayrallah, Culkin, Xia, Chen, Post, and
  Van~Durme}]{hu2019improvedparabank}
J.~Edward Hu, Huda Khayrallah, Ryan Culkin, Patrick Xia, Tongfei Chen, Matt
  Post, and Benjamin Van~Durme. 2019.
\newblock \href {https://doi.org/10.18653/v1/N19-1090} {Improved lexically
  constrained decoding for translation and monolingual rewriting}.
\newblock In \emph{Proceedings of the 2019 Conference of the North {A}merican
  Chapter of the Association for Computational Linguistics: Human Language
  Technologies, Volume 1 (Long and Short Papers)}, pages 839--850, Minneapolis,
  Minnesota. Association for Computational Linguistics.

\bibitem[{Iyyer et~al.(2018)Iyyer, Wieting, Gimpel, and
  Zettlemoyer}]{iyyer2018syntactic}
Mohit Iyyer, John Wieting, Kevin Gimpel, and Luke Zettlemoyer. 2018.
\newblock \href {https://doi.org/10.18653/v1/N18-1170} {Adversarial example
  generation with syntactically controlled paraphrase networks}.
\newblock In \emph{Proceedings of the 2018 Conference of the North {A}merican
  Chapter of the Association for Computational Linguistics: Human Language
  Technologies, Volume 1 (Long Papers)}, pages 1875--1885, New Orleans,
  Louisiana. Association for Computational Linguistics.

\bibitem[{Jia and Liang(2017)}]{jia2017adversarial}
Robin Jia and Percy Liang. 2017.
\newblock \href {https://doi.org/10.18653/v1/D17-1215} {Adversarial examples
  for evaluating reading comprehension systems}.
\newblock In \emph{Proceedings of the 2017 Conference on Empirical Methods in
  Natural Language Processing}, pages 2021--2031, Copenhagen, Denmark.
  Association for Computational Linguistics.

\bibitem[{Joshi et~al.(2017)Joshi, Choi, Weld, and
  Zettlemoyer}]{joshi2017triviaqa}
Mandar Joshi, Eunsol Choi, Daniel Weld, and Luke Zettlemoyer. 2017.
\newblock \href {https://doi.org/10.18653/v1/P17-1147} {{T}rivia{QA}: A large
  scale distantly supervised challenge dataset for reading comprehension}.
\newblock In \emph{Proceedings of the 55th Annual Meeting of the Association
  for Computational Linguistics (Volume 1: Long Papers)}, pages 1601--1611,
  Vancouver, Canada. Association for Computational Linguistics.

\bibitem[{Kincaid et~al.(1975)Kincaid, Fishburne~Jr, Rogers, and
  Chissom}]{kincaid1975derivation}
J~Peter Kincaid, Robert~P Fishburne~Jr, Richard~L Rogers, and Brad~S Chissom.
  1975.
\newblock Derivation of new readability formulas (automated readability index,
  fog count and flesch reading ease formula) for navy enlisted personnel.

\bibitem[{Lee et~al.(2019)Lee, Kim, and Hwang}]{lee2019qadiver}
Gyeongbok Lee, Sungdong Kim, and Seung-won Hwang. 2019.
\newblock Qadiver: Interactive framework for diagnosing qa models.
\newblock In \emph{Proceedings of the AAAI Conference on Artificial
  Intelligence}, volume~33, pages 9861--9862.

\bibitem[{Li et~al.(2018)Li, Ji, Du, Li, and Wang}]{li2018textbugger}
Jinfeng Li, Shouling Ji, Tianyu Du, Bao~Qin Li, and Ting Wang. 2018.
\newblock Textbugger: Generating adversarial text against real-world
  applications.
\newblock \emph{Network and Distributed System Security Symposium}.

\bibitem[{Lipton(2018)}]{lipton2016mythos}
Zachary~C. Lipton. 2018.
\newblock \href {https://doi.org/10.1145/3236386.3241340} {The mythos of model
  interpretability}.
\newblock \emph{{ACM} Queue}, 16(3):30.

\bibitem[{Liu and Stamm(2007)}]{liu2007fighting}
Changwei Liu and Sid Stamm. 2007.
\newblock Fighting unicode-obfuscated spam.
\newblock In \emph{Proceedings of the anti-phishing working groups 2nd annual
  eCrime researchers summit}, pages 45--59. ACM.

\bibitem[{Mudrakarta et~al.(2018)Mudrakarta, Taly, Sundararajan, and
  Dhamdhere}]{mudrakarta2018did}
Pramod~Kaushik Mudrakarta, Ankur Taly, Mukund Sundararajan, and Kedar
  Dhamdhere. 2018.
\newblock \href {https://doi.org/10.18653/v1/P18-1176} {Did the model
  understand the question?}
\newblock In \emph{Proceedings of the 56th Annual Meeting of the Association
  for Computational Linguistics (Volume 1: Long Papers)}, pages 1896--1906,
  Melbourne, Australia. Association for Computational Linguistics.

\bibitem[{Niu and Bansal(2018)}]{niu2018adversarial}
Tong Niu and Mohit Bansal. 2018.
\newblock \href {https://doi.org/10.18653/v1/K18-1047} {Adversarial
  over-sensitivity and over-stability strategies for dialogue models}.
\newblock In \emph{Proceedings of the 22nd Conference on Computational Natural
  Language Learning}, pages 486--496, Brussels, Belgium. Association for
  Computational Linguistics.

\bibitem[{Pearl(2019)}]{pearl2019seven}
Judea Pearl. 2019.
\newblock The seven tools of causal inference, with reflections on machine
  learning.
\newblock \emph{Commun. ACM}, 62(3):54--60.

\bibitem[{Pennington et~al.(2014)Pennington, Socher, and
  Manning}]{pennington2014glove}
Jeffrey Pennington, Richard Socher, and Christopher Manning. 2014.
\newblock \href {https://doi.org/10.3115/v1/D14-1162} {{G}love: Global vectors
  for word representation}.
\newblock In \emph{Proceedings of the 2014 Conference on Empirical Methods in
  Natural Language Processing ({EMNLP})}, pages 1532--1543, Doha, Qatar.
  Association for Computational Linguistics.

\bibitem[{Peters et~al.(2018)Peters, Neumann, Iyyer, Gardner, Clark, Lee, and
  Zettlemoyer}]{peters2018elmo}
Matthew Peters, Mark Neumann, Mohit Iyyer, Matt Gardner, Christopher Clark,
  Kenton Lee, and Luke Zettlemoyer. 2018.
\newblock \href {https://doi.org/10.18653/v1/N18-1202} {Deep contextualized
  word representations}.
\newblock In \emph{Proceedings of the 2018 Conference of the North {A}merican
  Chapter of the Association for Computational Linguistics: Human Language
  Technologies, Volume 1 (Long Papers)}, pages 2227--2237, New Orleans,
  Louisiana. Association for Computational Linguistics.

\bibitem[{Poursabzi-Sangdeh et~al.(2018)Poursabzi-Sangdeh, Goldstein, Hofman,
  Vaughan, and Wallach}]{poursabzi2018manipulating}
Forough Poursabzi-Sangdeh, Daniel~G Goldstein, Jake~M Hofman, Jennifer~Wortman
  Vaughan, and Hanna Wallach. 2018.
\newblock Manipulating and measuring model interpretability.
\newblock \emph{arXiv preprint arXiv:1802.07810}.

\bibitem[{Rajpurkar et~al.(2016)Rajpurkar, Zhang, Lopyrev, and
  Liang}]{rajpurkar2016squad}
Pranav Rajpurkar, Jian Zhang, Konstantin Lopyrev, and Percy Liang. 2016.
\newblock \href {https://doi.org/10.18653/v1/D16-1264} {{SQ}u{AD}: 100,000+
  questions for machine comprehension of text}.
\newblock In \emph{Proceedings of the 2016 Conference on Empirical Methods in
  Natural Language Processing}, pages 2383--2392, Austin, Texas. Association
  for Computational Linguistics.

\bibitem[{Ribeiro et~al.(2016)Ribeiro, Singh, and Guestrin}]{ribeiro2016should}
Marco~Tulio Ribeiro, Sameer Singh, and Carlos Guestrin. 2016.
\newblock Why should i trust you?: Explaining the predictions of any
  classifier.
\newblock In \emph{Proceedings of the 22nd ACM SIGKDD international conference
  on knowledge discovery and data mining}, pages 1135--1144. ACM.

\bibitem[{Ribeiro et~al.(2018)Ribeiro, Singh, and Guestrin}]{ribeiro2018sears}
Marco~Tulio Ribeiro, Sameer Singh, and Carlos Guestrin. 2018.
\newblock \href {https://doi.org/10.18653/v1/P18-1079} {Semantically equivalent
  adversarial rules for debugging {NLP} models}.
\newblock In \emph{Proceedings of the 56th Annual Meeting of the Association
  for Computational Linguistics (Volume 1: Long Papers)}, pages 856--865,
  Melbourne, Australia. Association for Computational Linguistics.

\bibitem[{Seo et~al.(2017)Seo, Kembhavi, Farhadi, and
  Hajishirzi}]{seo2016bidaf}
Min~Joon Seo, Aniruddha Kembhavi, Ali Farhadi, and Hannaneh Hajishirzi. 2017.
\newblock \href {https://openreview.net/forum?id=HJ0UKP9ge} {Bidirectional
  attention flow for machine comprehension}.
\newblock In \emph{5th International Conference on Learning Representations,
  {ICLR} 2017, Toulon, France, April 24-26, 2017, Conference Track
  Proceedings}.

\bibitem[{Tram{\`{e}}r et~al.(2018)Tram{\`{e}}r, Kurakin, Papernot, Goodfellow,
  Boneh, and McDaniel}]{tramer2017ensemble}
Florian Tram{\`{e}}r, Alexey Kurakin, Nicolas Papernot, Ian~J. Goodfellow, Dan
  Boneh, and Patrick~D. McDaniel. 2018.
\newblock \href {https://openreview.net/forum?id=rkZvSe-RZ} {Ensemble
  adversarial training: Attacks and defenses}.
\newblock In \emph{6th International Conference on Learning Representations,
  {ICLR} 2018, Vancouver, BC, Canada, April 30 - May 3, 2018, Conference Track
  Proceedings}.

\bibitem[{Vaswani et~al.(2017)Vaswani, Shazeer, Parmar, Uszkoreit, Jones,
  Gomez, Kaiser, and Polosukhin}]{vaswani2017attention}
Ashish Vaswani, Noam Shazeer, Niki Parmar, Jakob Uszkoreit, Llion Jones,
  Aidan~N Gomez, {\L}ukasz Kaiser, and Illia Polosukhin. 2017.
\newblock Attention is all you need.
\newblock In \emph{Advances in neural information processing systems}, pages
  5998--6008.

\bibitem[{Wu et~al.(2019)Wu, Ribeiro, Heer, and Weld}]{wu2019errudite}
Tongshuang Wu, Marco~Tulio Ribeiro, Jeffrey Heer, and Daniel~S Weld. 2019.
\newblock Errudite: Scalable, reproducible, and testable error analysis.
\newblock In \emph{Proceedings of the 57th Conference of the Association for
  Computational Linguistics}, pages 747--763.

\bibitem[{Zhang et~al.(2019)Zhang, Sheng, and Alhazmi}]{zhang2019survey}
Wei~Emma Zhang, Quan~Z. Sheng, and Ahoud Abdulrahmn~F. Alhazmi. 2019.
\newblock \href {http://arxiv.org/abs/1901.06796} {Generating textual
  adversarial examples for deep learning models: {A} survey}.
\newblock \emph{CoRR}, abs/1901.06796.

\bibitem[{Zhao et~al.(2018)Zhao, Dua, and Singh}]{zhao2017generating}
Zhengli Zhao, Dheeru Dua, and Sameer Singh. 2018.
\newblock \href {https://openreview.net/forum?id=H1BLjgZCb} {Generating natural
  adversarial examples}.
\newblock In \emph{6th International Conference on Learning Representations,
  {ICLR} 2018, Vancouver, BC, Canada, April 30 - May 3, 2018, Conference Track
  Proceedings}.

\end{thebibliography}
\bibliographystyle{acl_natbib}

\end{document}